\documentclass[11pt]{article}

\usepackage[preprint]{acl}

\usepackage{times}
\usepackage{latexsym}
\usepackage[T1]{fontenc}
\usepackage[utf8]{inputenc}
\usepackage{microtype}
\usepackage{inconsolata}

\usepackage{graphicx}
\usepackage{booktabs}
\usepackage{colortbl}
\usepackage{xcolor}
\usepackage{adjustbox}
\usepackage{multirow}
\usepackage{enumitem}
\usepackage{subcaption}
\usepackage{amsmath}
\usepackage{amsfonts}
\usepackage{arydshln}
\usepackage[most]{tcolorbox}
\usepackage{fvextra}
\usepackage{algorithm}
\usepackage{algorithmic}
\usepackage{cuted}
\usetikzlibrary{positioning}

\definecolor{lightblue}{rgb}{0.85,0.92,1}
\definecolor{lightyellow}{rgb}{1,0.95,0.8}
\definecolor{darkblue}{rgb}{0,0,0.5}
\hypersetup{colorlinks=true, citecolor=darkblue, linkcolor=darkblue, urlcolor=darkblue}

\title{Dynamic Dual-Granularity Skill Bank for Agentic RL}

\author{Songjun Tu$^{1,2}$, Chengdong Xu$^{3,2}$, Qichao Zhang$^{1}$, Yaocheng Zhang$^{1}$, \\
\textbf{Xiangyuan Lan}$^{2}$, \textbf{Linjing Li}$^{1}$, \textbf{Dong Li}$^{4}$, \textbf{Dongbin Zhao}$^{1,2}$ \\
Institute of Automation, Chinese Academy of Sciences$^1$\\
Pengcheng Laboratory$^2$ \quad Sun Yat-Sen University$^3$ \quad MemoraX AI$^4$}

\begin{document}
\maketitle
\begin{abstract}
    Agentic RL can benefit substantially from reusable experience, but existing skill-based methods mostly focus on trajectory-level guidance and provide limited support for the principled evaluation and maintenance of an evolving skill memory.
    We propose \textbf{D2Skill}, a \textbf{dynamic dual-granularity skill bank} for agentic RL that organizes reusable experience into \textbf{task skills} for high-level guidance and \textbf{step skills} for fine-grained decision support and error correction. D2Skill jointly trains the policy and skill bank through paired baseline and skill-injected rollouts under the same policy, using their performance gap to derive hindsight utility signals for both skill updating and policy optimization. Built entirely from training-time experience, the skill bank is continuously expanded through reflection and maintained with utility-aware retrieval and pruning. Experiments on \textsc{ALFWorld}, \textsc{WebShop}, and \textsc{Search-Augmented QA} tasks show that D2Skill substantially improves performance over skill-free baselines across models of different scales. Further ablations and analyses show that both dual-granularity skill modeling and dynamic skill maintenance are critical to these gains, while the learned skills exhibit higher utility, transfer across evaluation settings, and introduce only modest training overhead.
\textbf{Code Page:} \href{https://github.com/TU2021/D2Skill-AgenticRL}{https://github.com/TU2021/D2Skill-AgenticRL}.
\end{abstract}

\begin{figure}[t]
    \centering
    \begin{subfigure}[t]{0.98\columnwidth}
        \centering
        \includegraphics[width=\textwidth]{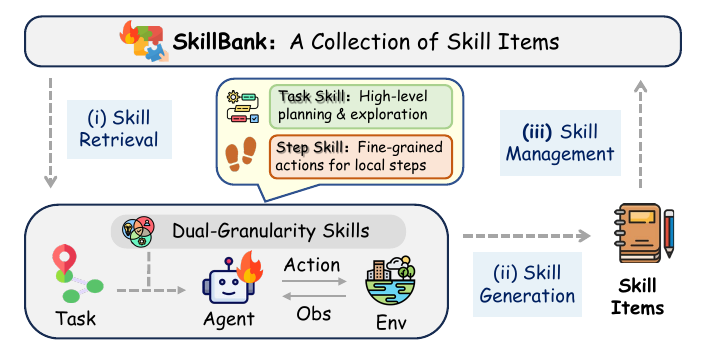}
        \caption{D2Skill Framework}
        \label{fig:overview_4panel_sub1}
    \end{subfigure}

    \vspace{0.35em}

    \begin{subfigure}[t]{0.98\columnwidth}
        \centering
        \includegraphics[width=\textwidth]{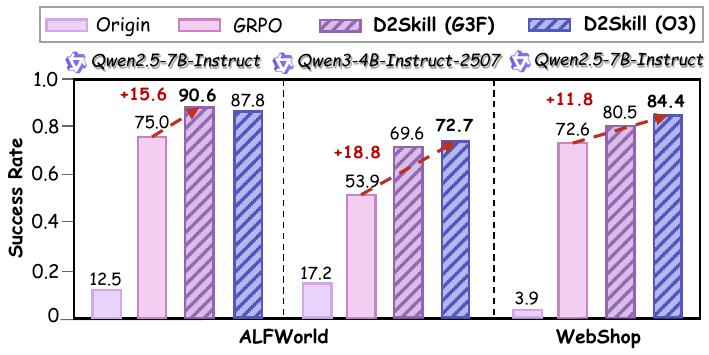}
        \caption{Main Results}
        \label{fig:overview_4panel_sub2}
    \end{subfigure}
    \caption{\textbf{Overview of D2Skill.}
    (a) The dynamic dual-granularity skill bank with retrieval, reflection-driven generation, and management.
    (b) Overall results on \textsc{ALFWorld} and \textsc{WebShop}.}
    \label{fig:overview_4panel}
\end{figure}

\section{Introduction}
Agentic reinforcement learning (RL) has emerged as a promising paradigm for training language-based agents on long-horizon decision-making tasks, including interactive environments \citep{jiang2026wovr}, web search \citep{zhang2025criticsearch}, and research scenarios \citep{tu2026paperaudit}. In these settings, agents act through textual interfaces based on task descriptions and limited interaction history, making the problem severely partially observable and credit assignment difficult over long horizons \citep{zhang2025landscape}. Under sparse rewards and large action spaces, effective learning therefore requires reusable knowledge that can transfer across tasks \citep{feng2025group}.

Recent work addresses these challenges by introducing additional supervision signals for agentic RL.
Some methods use outcome-based credit assignment to provide process rewards~\citep{feng2025group}, while others derive hindsight supervision from completed trajectories~\citep{yu2025memagent}.
Another line of work enables agents to accumulate and refine experience across tasks during training~\citep{zhai2025agentevolver,cai2025flex,cai2025training}.
Within this line, reusable skills have emerged as an effective abstraction of past experience and have shown strong empirical gains in agentic RL~\citep{wang2026openclaw}.
For example, SkillRL~\citep{xia2026skillrl} builds a skill bank from past trajectories and retrieves relevant skills to guide policy interaction, improving exploration efficiency in long-horizon tasks.

However, existing skill-based and reflection-driven frameworks remain limited in two respects.
First, many existing methods derive skills from complete trajectories and emphasize task-level reflection, which provides high-level guidance but is less effective for correcting fine-grained errors at individual interaction steps~\citep{xia2026skillrl,zhang2026memrl}.
Second, as training progresses, the skill bank grows continuously, making retrieval increasingly sensitive to redundancy and skill quality.
Without principled mechanisms for skill evaluation and pruning, ineffective or redundant skills can degrade retrieved guidance and hinder policy optimization~\citep{zhou2025memento,zhou2026memento}.

To address these limitations, we propose \textbf{D}ynamic \textbf{D}ual-Granularity \textbf{Skill} Bank (\textbf{D2Skill}) for agentic RL, a framework that maintains reusable skills at both the task and step granularities throughout training. As illustrated in Figure~\ref{fig:overview_4panel_sub1}, D2Skill combines task skills for high-level guidance with step skills for local decision support, and contrasts skill-injected and non-injected rollouts under the same policy to estimate hindsight skill utility for retrieval, maintenance, and policy optimization. Figure~\ref{fig:overview_4panel_sub2} previews the main benchmark result: on \textsc{ALFWorld} and \textsc{WebShop}, D2Skill consistently outperforms strong skill-free and skill-augmented baselines. It also generalizes favorably to \textsc{Search-Augmented QA}, while ablations validate the roles of dual-granularity skills and dynamic memory management.

The main contributions are as follows:
\begin{tcolorbox}[breakable,
    enhanced,
    colback=black!2,
    colframe=black!15,
    boxrule=0.1pt,
    arc=1mm,
    sharp corners=southwest,
    sharp corners=southeast,
    left=0.0mm,right=1.0mm,top=1.0mm,bottom=1.0mm
]
\begin{enumerate}[leftmargin=2em, itemsep=2pt, topsep=2pt]
\item We present \textbf{D2Skill}, a dynamic dual-granularity skill bank framework for agentic RL that organizes reusable experience into \emph{task skills} for high-level guidance and \emph{step skills} for local interaction support.
\item We develop a \textbf{joint training paradigm} in which the policy and skill bank co-evolve: skills are expanded through \textbf{reflection} and maintained with \textbf{utility-aware retrieval and pruning}, enabling more effective and efficient memory use throughout training.
\item We evaluate D2Skill on \textsc{ALFWorld}, \textsc{WebShop}, and \textsc{Search-Augmented QA} using \textsc{Qwen2.5-7B-Instruct} and \textsc{Qwen3-4B-Instruct}. D2Skill \textbf{consistently outperforms skill-free baselines}, with ablations validating its core design.
\end{enumerate}
\end{tcolorbox}

\section{Motivation}
\label{sec:motivation}

\begin{figure}[t]
    \centering
    \includegraphics[width=\columnwidth]{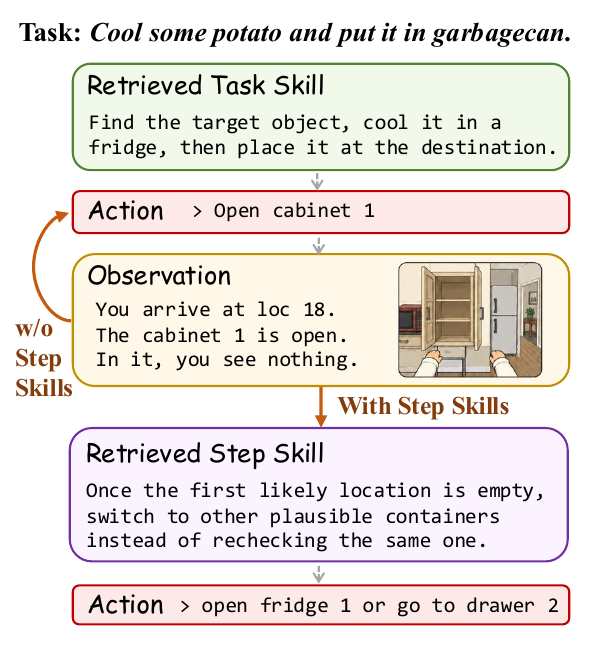}
    \caption{\textbf{Step skills correct local search failures.} Without step skills, the agent can repeat the same failed search after a negative observation. A retrieved step skill redirects it to another plausible location.}
    \label{fig:step_skill_motivation}
\end{figure}

\begin{figure*}[t]
    \centering
    \begin{subfigure}[t]{0.48\textwidth}
        \centering
        \includegraphics[width=\textwidth]{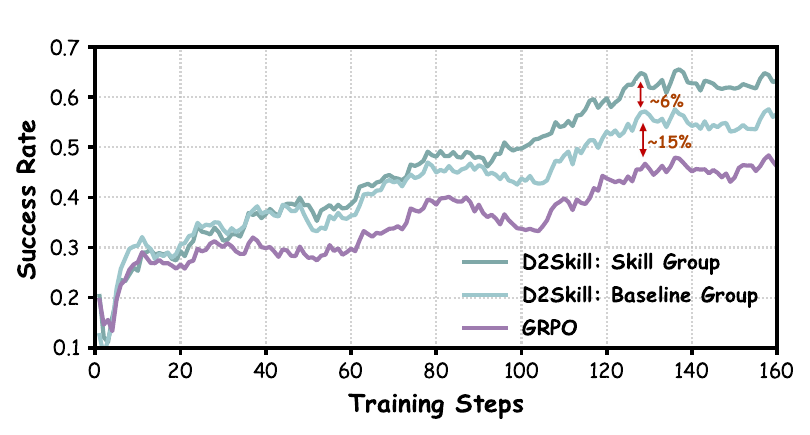}
        \caption{Training Curves of Success Rate}
        \label{fig:overview_4panel_sub3}
    \end{subfigure}
    \hfill
    \begin{subfigure}[t]{0.48\textwidth}
        \centering
        \includegraphics[width=\textwidth]{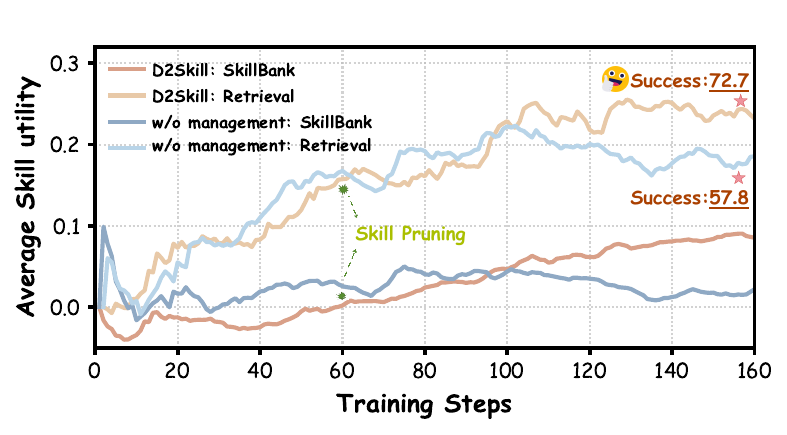}
        \caption{Skill Bank Dynamics of Utility}
        \label{fig:overview_4panel_sub4}
    \end{subfigure}
    \caption{\textbf{Training and skill-bank dynamics motivating D2Skill.}
    (a) \textsc{ALFWorld} training curves for the D2Skill skill group, the paired baseline group, and \textsc{GRPO}.
    (b) Skill-bank dynamics with and without management, measured by average skill utility and retrieval statistics.}
    \label{fig:motivation_panels}
\end{figure*}

\subsection{Local Errors Need Fine-Grained Skills}

A limitation of task-level skills is that they mainly provide global guidance, while many failures in long-horizon agentic interaction arise from \emph{local} errors after the overall plan is already reasonable. Once the agent reaches a concrete decision point, it must react to the current observation rather than rely only on a trajectory-level plan. Figure~\ref{fig:step_skill_motivation} illustrates this gap: a task skill can correctly prescribe the high-level workflow for cooling an object and placing it at the destination, yet it does not specify how to recover when the agent opens a plausible container and finds it empty. What is missing at that point is a short, reusable rule for local correction, such as switching to another plausible location instead of repeating the same failed search. This motivates introducing \textbf{step skills}, which provide observation-conditioned, fine-grained corrective guidance that complements task-level planning. We focus on two levels of abstraction because they map naturally to the two dominant failure modes in agentic RL: global planning and local recovery.

\subsection{Growing Skill Banks Benefit from Dynamic Management}

A limitation of static skill-bank reuse is that the memory keeps growing during training, while not all generated skills remain equally useful. Some skills are overly specific to a single failure, some become redundant as stronger skills are acquired later, and some continue to occupy retrieval and prompt budget without improving decisions. This issue is especially relevant to methods such as SkillRL, which retrieve stored memories but do not explicitly estimate their utility during training. As a result, simply accumulating more skills does not guarantee better guidance. Figure~\ref{fig:overview_4panel_sub3} shows that retrieved skills can provide measurable benefit, while Figure~\ref{fig:overview_4panel_sub4} shows that utility-aware management preserves higher-quality retrieved skills over time. These observations motivate \textbf{dynamic skill management}: instead of treating the bank as a static memory, D2Skill explicitly ranks, retains, and prunes skills according to the estimated utility.

\section{Method}

\subsection{Overall Framework}

\begin{figure*}[t]
\centering
\includegraphics[width=\textwidth]{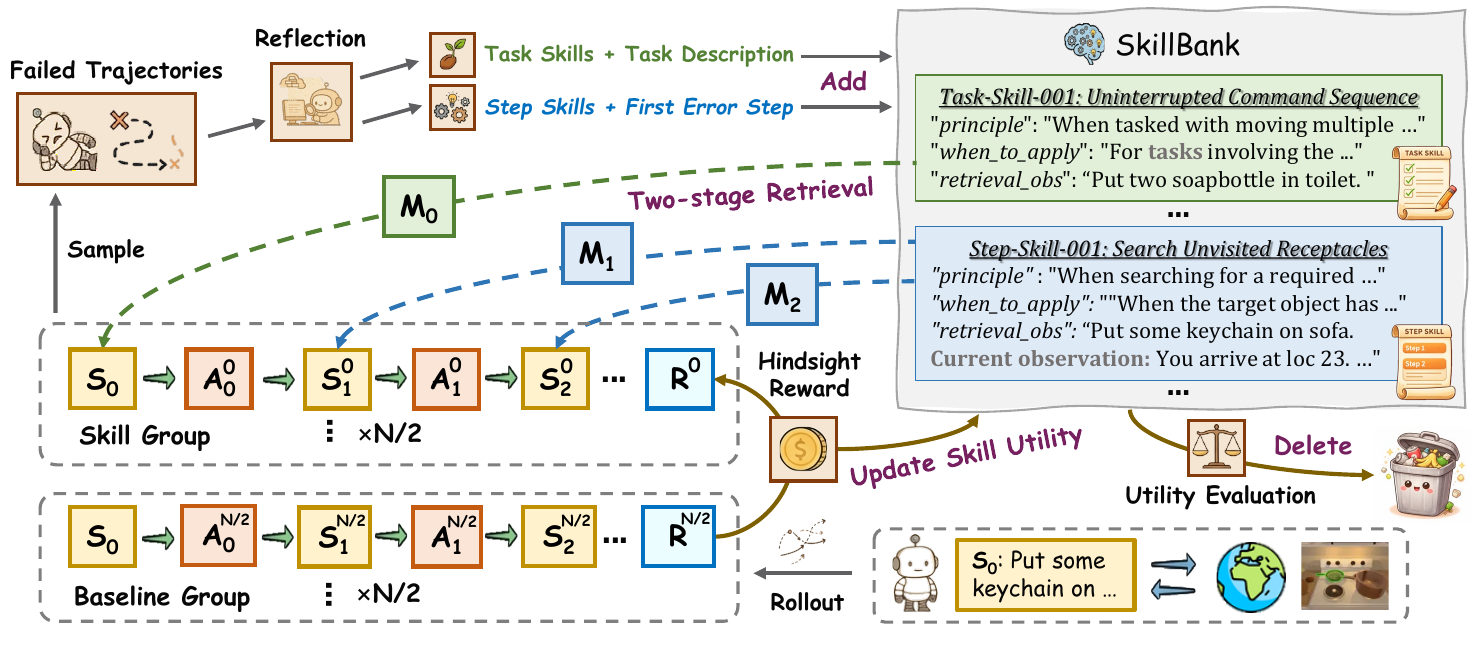}
\caption{\textbf{Overall framework of D2Skill.} D2Skill couples RL with a dynamic dual-granularity skill bank. For each task, training rollouts are divided into a baseline group and a skill group, whose performance gap yields hindsight signals for policy optimization and skill utility estimation. When performance is poor, reflection on representative failed trajectories produces \emph{task skills} for high-level guidance and \emph{step skills} for local error correction. Skills are stored with retrieval keys, reused during interaction, and periodically pruned by utility-based bank management.}
\label{fig_main}
\vspace{-0.5em}
\end{figure*}

D2Skill augments RL with a dynamic skill bank, as illustrated in Fig.~\ref{fig_main}. The framework has three tightly coupled parts.

\paragraph{RL training with skill injection.}
Each training group contains paired baseline and skill-injected rollouts under the same policy. Their performance gap provides hindsight signals for policy optimization and skill utility estimation (Section~\ref{sec_method_1}).

\paragraph{Reflection-driven skill generation.}
When performance is poor, a reflection module summarizes representative trajectories into reusable \emph{task skills} and \emph{step skills}, which are normalized and added to the bank.

\paragraph{Skill retrieval and bank management.}
During interaction, relevant skills are retrieved from the bank and injected into the policy context. Skill utilities are updated online, and the bank is periodically pruned to maintain a compact and effective memory (Section~\ref{sec_method_2}).

\subsection{RL Training with Skill Injection and Hindsight Optimization}
\label{sec_method_1}

\paragraph{Rollout with skill injection.}
For each task $g$, we sample a group of $N$ parallel trajectories, denoted by $\mathcal{G}_g$.
The group is evenly divided into a \emph{\textbf{skill group}} $\mathcal{G}_g^{\mathrm{skill}}$
and a \emph{\textbf{baseline group}} $\mathcal{G}_g^{\mathrm{base}}$, each containing $N/2$ trajectories.
Let $b_i \in \{0,1\}$ denote the group indicator for trajectory $i \in \mathcal{G}_g$,
where $b_i=1$ indicates $i \in \mathcal{G}_g^{\mathrm{skill}}$ and $b_i=0$ indicates
$i \in \mathcal{G}_g^{\mathrm{base}}$.
Trajectories in the skill group retrieve skills from the skill bank during interaction,
while those in the baseline group follow the same policy without skill injection.
Let $Y_i \in \{0,1\}$ denote the terminal success indicator of trajectory $i$. For each task $g$, the baseline success rate and skill-group success rate are
$\bar{Y}_g^{\mathrm{base}}=\tfrac{1}{|\mathcal{G}_g^{\mathrm{base}}|}\sum_{i \in \mathcal{G}_g^{\mathrm{base}}} Y_i$
and
$\bar{Y}_g^{\mathrm{skill}}=\tfrac{1}{|\mathcal{G}_g^{\mathrm{skill}}|}\sum_{i \in \mathcal{G}_g^{\mathrm{skill}}} Y_i$.

\paragraph{Hindsight signals and utility updates.}
We use the performance gap between the skill group and the baseline group
to construct hindsight signals for updating skill utilities.
For each task $g$, the task-level hindsight signal $\Delta_g^{\mathrm{task}}$
and the trajectory-level credit $c_i$ for step skills retrieved along
skill-injected trajectory $i$ are defined as
\begin{equation}
\Delta_g^{\mathrm{task}}=\bar{Y}_g^{\mathrm{skill}}-\bar{Y}_g^{\mathrm{base}},
\quad
c_i=Y_i - \bar{Y}_g^{\mathrm{base}}.
\end{equation}
Each skill $m$ maintains a utility $u_m$ updated using an exponential moving average.
For a given task $g$, all retrieved task skills share the same signal
$\Delta_g^{\mathrm{task}}$, since the task context is identical for the whole group.
In contrast, multiple step skills may be retrieved at different steps and
from different trajectories, and each retrieved step skill is updated using
the credit of the trajectory in which it appears.
The updates are defined as
\begin{equation}
\begin{aligned}
u_m &\leftarrow (1-\beta_{\mathrm{task}})u_m
      +\beta_{\mathrm{task}}\Delta_g^{\mathrm{task}},\\
u_m &\leftarrow (1-\beta_{\mathrm{step}})u_m
      +\beta_{\mathrm{step}}c_i,
\end{aligned}
\end{equation}
where the first rule is applied to task skills retrieved in task $g$,
and the second rule is applied to each step skill retrieved along
skill-injected trajectory $i$.

\paragraph{Hindsight intrinsic reward shaping.}
To encourage effective use of retrieved skills, we introduce a hindsight intrinsic reward
for trajectories in the skill group.
For each skill-injected trajectory $i \in \mathcal{G}_g^{\mathrm{skill}}$,
the hindsight intrinsic reward is defined as
\begin{equation}
R_i^{\mathrm{int}}=\lambda\bigl(Y_i - \bar{Y}_g^{\mathrm{base}}\bigr),
\end{equation}
where $\lambda$ controls the strength of the shaping signal.
This term measures performance gain over the baseline and encourages effective skill usage.
The hindsight intrinsic reward is applied at the end of each skill-injected trajectory and included in the policy optimization.

\paragraph{Policy optimization with skill-augmented returns.}
The policy is optimized on the full samples.
For each task $g$, trajectories in the skill group $\mathcal{G}_g^{\mathrm{skill}}$
are generated under skill-augmented context and receive an additional reward $R^{\mathrm{int}}$.
Let $R_i$ denote the origin return of trajectory $i$.
For skill-injected trajectories, the return is augmented with $R_i^{\mathrm{int}}$, and advantages are computed by group normalization over the whole trajectory group:
\begin{equation}
\begin{aligned}
\tilde{R}_i &=
\begin{cases}
R_i + R_i^{\mathrm{int}}, & i \in \mathcal{G}_g^{\mathrm{skill}}, \\
R_i, & i \in \mathcal{G}_g^{\mathrm{base}},
\end{cases}\\
A_i &=
\frac{\tilde{R}_i-\operatorname{mean}(\{\tilde{R}_j\}_{j\in\mathcal{G}_g})}
{\operatorname{std}(\{\tilde{R}_j\}_{j\in\mathcal{G}_g})}.
\end{aligned}
\end{equation}
We optimize the policy with a GRPO-style clipped objective using the hindsight-augmented advantages above; the full loss is given in Appendix~\ref{app:skill_bank}.

\subsection{Skill Generation, Retrieval, and Bank Management}
\label{sec_method_2}
\paragraph{Reflection and skill generation.} 
Reflection is triggered only for task groups with low performance, i.e., when $\bar{Y}_g^{\mathrm{skill}} < \tau_{\mathrm{ref}}$, where $\tau_{\mathrm{ref}}$ is a reflection threshold. For each such task $g$, we sample one failed trajectory $\tau_g^{-}$ from the skill group and, \textbf{when available, one successful trajectory $\tau_g^{+}$ from either the skill or the baseline group, and use them for skill generation}. The reflector produces at most one task skill and one step skill for each task group, formalized as
\begin{equation}
\begin{aligned}
m_g^{\mathrm{task}}
&= f^{\mathrm{task}}_{\mathrm{reflect}}(g,\tau_g^{-},\tau_g^{+}),\\
(m_g^{\mathrm{step}}, o_j)
&= f^{\mathrm{step}}_{\mathrm{reflect}}(g,\tau_g^{-},\tau_g^{+}),
\end{aligned}
\end{equation}
where $f_{\mathrm{reflect}}$ denotes an external reflector LLM used for skill generation, and $o_j$ denotes the observation at the earliest failure step $j$ identified from the sampled failed trajectory.

For each skill $m$, we define a retrieval key $k_m$ that determines when the skill is applicable. For $m \in \mathcal{M}_{\mathrm{task}}$, the key is defined as $k_m = g$. For $m \in \mathcal{M}_{\mathrm{step}}$, the key is defined as $k_m = (g, o_j)$. New skills are inserted into the skill bank after deduplication and participate in subsequent retrieval and utility updates.

\paragraph{Two-stage skill retrieval.}
When interacting with environment, skills are retrieved from the skill bank
by matching the current query key with the retrieval key $k_m$ of each skill.
For task-level retrieval, the query key is $q = g$,
while for step-level retrieval the query key is $q_t = (g, o_t)$,
where $g$ denotes the task identifier and $o_t$ is the observation at step $t$.

In the first stage, we retrieve the \textbf{\textit{top-m}} candidate skills
from the pool $\mathcal{M} \in \{\mathcal{M}_{\mathrm{task}}, \mathcal{M}_{\mathrm{step}}\}$
according to cosine similarity between the embedding of $q$ and $k_m$.
A minimum similarity threshold $\tau_{\mathrm{sim}}$ is applied,
and only skills satisfying $\mathrm{sim}(q,k_m)\ge\tau_{\mathrm{sim}}$
are retained.

In the second stage, the candidates are ranked using a combination of
semantic similarity and utility-based exploration.
For each skill $m \in \mathcal{M}$, we define the selection score
\begin{equation}
\begin{split}
&\mathrm{score}(m)
=\ \alpha\,\widehat{\mathrm{sim}}(m,q)\\
&+(1-\alpha)
\left(u_m+\eta
\sqrt{\frac{\log(1+N_r)}{1+n_m}}\right),
\end{split}
\label{eq-score}
\end{equation}

where $\widehat{\mathrm{sim}}(m,q)\in[0,1]$ is the normalized cosine similarity,
$u_m$ is the utility of skill $m$,
$n_m$ is the number of times the skill has been retrieved, and
$N_r=\sum_{m' \in \mathcal{M}} n_{m'}$
is the total retrieval count in the active pool.
The second term corresponds to a UCB-style bonus that encourages exploration of skills with low retrieval counts.
The \textbf{\textit{top-k}} ($<$top-m) skills ranked by this score are injected into the policy context.

\paragraph{Skill pruning by utility.}
To prevent unbounded growth of the skill bank,
we periodically prune each skill pool
$\mathcal{M}$
after validation intervals.
Each pool is associated with a capacity limit $N_{\max}$.
If
\(|\mathcal{M}| > N_{\max},\)
each skill $m \in \mathcal{M}$ is assigned an eviction score
\begin{equation}
\mathrm{evict}(m)=u_m+\eta\sqrt{\frac{\log(1+N_r)}{1+n_m}}.
\label{eq-evict}
\end{equation}

Then, skills are sorted by $\mathrm{evict}(m)$ in ascending order,
and the lowest-scoring ones are removed until
$|\mathcal{M}| \le N_{\max}$.
Skills created within the last $T_{\mathrm{prot}}$ training steps,
i.e., $t - t_m^{\mathrm{create}} < T_{\mathrm{prot}}$,
are excluded from eviction to allow sufficient evaluation.


\section{Experiments}

We evaluate D2Skill on three representative agentic settings: the interactive benchmarks \textsc{ALFWorld} \citep{shridhar2020alfworld} and \textsc{WebShop} \citep{yao2022webshop}, and \textsc{Search-Augmented QA} \citep{jin2025searchr1}. Across these settings, we compare D2Skill against skill-free RL baselines and prior methods that augment agents with memory, skills, or search. Detailed training protocols and baseline descriptions are deferred to Appendix~\ref{app:exp-details}.
Our experiments address three questions:
\begin{tcolorbox}[breakable,
    enhanced,
    colback=black!2,
    colframe=black!15,
    boxrule=0.35pt,
    arc=1.5mm,
    sharp corners=southwest,
    sharp corners=southeast,
    left=1.2mm,right=1.2mm,top=1.2mm,bottom=1.2mm
]
\begin{enumerate}[leftmargin=2em, itemsep=2pt, topsep=2pt]
\item \textbf{Main Performance:} Does D2Skill outperform standard RL and existing baselines on agentic tasks? (Section~\ref{exp:main})
\item \textbf{Ablation:} What is the contribution of each major component to the overall gains? (Section~\ref{exp:ablation}) 
\item \textbf{Analysis:} How does the dynamic skill bank affect skill utility, training dynamics, computational overhead, and policy generalization without skill-bank access at evaluation time? (Section~\ref{exp:analysis})
\end{enumerate}
\end{tcolorbox}

\subsection{Main Performance}
\label{exp:main}
\begin{table*}[ht]
\setlength{\tabcolsep}{5pt}
\centering
\small
\caption{Performance on \textsc{ALFWorld} and \textsc{WebShop}. 
For \textsc{ALFWorld}, we report the average success rate (\%) for each subtask and the overall success rate. 
For \textsc{WebShop}, we report the average score and average success rate (\%). 
Evaluation details are given in Appendix~\ref{app:benchmarks}. 
$^*$ denotes results replicated from~\citep{feng2025group} and \citep{xia2026skillrl}. 
The best and second-best results are highlighted in \textcolor{red!50}{red} and \textcolor{blue!50}{blue}, respectively.}
\label{tab:performance}
\resizebox{\textwidth}{!}{
\begin{tabular}{lccccccc|cc}
\toprule
\multirow{2}{*}{\textbf{Method}} & \multicolumn{7}{c|}{\textbf{ALFWorld}} & \multicolumn{2}{c}{\textbf{WebShop}} \\
 & Pick & Clean & Cool & Look & Heat & Pick2 & All & Score & Success \\
\midrule
\rowcolor{gray!10} \multicolumn{10}{l}{Closed-source LLMs} \\
\textit{Gemini-3-Flash} & 96.4 & 57.1 & 96.2 & 85.7 & 72.2 & 95.3 & 85.2 & 14.1 & 16.5  \\
\textit{O3} & 64.3 & 19.1 & 23.1 & 64.3 & 33.3 & 61.9 & 43.8 & 5.8 & 4.7  \\
\midrule
\rowcolor{gray!10} \multicolumn{10}{l}{Base Model: \textit{Qwen2.5-7B-Instruct}} \\
Origin & 17.9 & 4.8 & 3.8 & 64.3 & 0.0 & 5.3 & 12.5 & 16.6 & 3.9 \\
GRPO & 88.3 & 73.3 & 76.0 & \cellcolor{blue!8}83.3 & 81.3 & 40.0 & 75.0 & 86.0 & 72.6 \\
Mem0+GRPO* & 78.1 & 56.1 & 65.0 & 54.8 & 31.0 & 26.9 & 54.7 & 58.1 & 37.5  \\
SimpleMem+GRPO* & 89.5 & 60.0 & 64.9 & 36.3 & 50.0 & 26.3 & 62.5 & 67.8 & 46.9 \\
SkillRL(\textit{O3})* & \cellcolor{blue!8}94.3 & 90.6 & \cellcolor{blue!8}92.0 & \cellcolor{blue!8}83.3 & 93.7 & \cellcolor{red!8}80.0 & \cellcolor{blue!8}89.1 & 85.2 & 72.7 \\
\textbf{D2Skill}(\textit{Gemini-3-Flash}) & \cellcolor{red!8}97.1 & \cellcolor{red!8}100.0 & 75.0 & \cellcolor{red!8}87.5 & \cellcolor{red!8}100.0 & \cellcolor{blue!8}78.6 & \cellcolor{red!8}90.6 & \cellcolor{red!8}91.1 & \cellcolor{blue!8}80.5  \\
\textbf{D2Skill}(\textit{O3}) & 93.8 & \cellcolor{blue!8}94.7 & \cellcolor{red!8}95.5 & 77.8 & \cellcolor{blue!8}95.0 & 72.0 & 87.8 & \cellcolor{blue!8}90.1 & \cellcolor{red!8}84.4 \\
\textbf{D2Skill}(\textit{Self}) & \cellcolor{blue!8}94.3 & 93.8 & 73.1 & 75.7 & 75.0 & 65.0 & 82.8 & 89.9 & \cellcolor{blue!8}81.3 \\
\midrule
\rowcolor{gray!10} \multicolumn{10}{l}{Base Model: \textit{Qwen3-4B-Instruct-2507}} \\
Origin & 50.0 & 9.5 & 0.0 & 2.1 & 11.1 & 4.8 & 17.2 & 0.0 & 0.0  \\
GRPO & 73.5 & 46.6 & 48.0 & 61.1 & 62.5 & 20.0 & 53.9 & 87.0 & 74.2 \\
SkillRL(\textit{O3}) & \cellcolor{red!8}90.0 & \cellcolor{red!8}92.3 & 52.0 & \cellcolor{blue!8}63.6 & 42.9 & 40.9 & 67.2 & 89.4 & 75.7  \\
\textbf{D2Skill}(\textit{Gemini-3-Flash}) & 88.6 & 75.0 & 54.2 & \cellcolor{red!8}66.7 & \cellcolor{blue!8}60.0 & \cellcolor{red!8}52.6 & \cellcolor{blue!8}69.6 & \cellcolor{red!8}90.0 & \cellcolor{red!8}82.8 \\
\textbf{D2Skill}(\textit{O3}) & \cellcolor{blue!8}89.4 & 72.4 & \cellcolor{red!8}66.7 & 54.5 & \cellcolor{blue!8}60.0 & \cellcolor{blue!8}50.0 & \cellcolor{red!8}72.7 & 89.1 & 75.8 \\
\textbf{D2Skill}(\textit{Self}) & 85.7 & \cellcolor{blue!8}79.2 & \cellcolor{blue!8}57.7 & 60.0 & \cellcolor{red!8}73.3 & 35.0 & 62.5 & \cellcolor{blue!8}89.6 & \cellcolor{blue!8}77.4 \\
\midrule
\rowcolor{gray!10} \multicolumn{10}{l}{Base Model: \textit{Qwen3-4B-Instruct-2507 + SFT}} \\
Origin & 53.6 & 28.6 & 46.2 & 71.4 & 55.5 & 38.1 & 47.7 & 65.6  & 53.1 \\
GRPO(\textit{40-Steps}) & 89.7 & 77.8 & 85.7 & \cellcolor{red!8}91.6 & 86.7 & 69.6 & 83.6 & 77.4 & 67.2 \\
GRPO(\textit{120-Steps}) & \cellcolor{red!8}100.0 & 95.2 & 80.8 & \cellcolor{blue!8}88.9 & 78.6 & \cellcolor{blue!8}88.3 & \cellcolor{blue!8}92.9 & \cellcolor{blue!8}88.2 & \cellcolor{blue!8}79.9 \\
\textbf{D2Skill}(\textit{40-Steps}) & 92.9 & \cellcolor{red!8}100.0 & \cellcolor{blue!8}95.2 & 80.0 & \cellcolor{red!8}90.9 & 86.7 & 92.2 & 84.1 & 71.9 \\
\textbf{D2Skill}(\textit{120-Steps}) & \cellcolor{blue!8}97.6 & \cellcolor{blue!8}95.8 & \cellcolor{red!8}100.0 & \cellcolor{blue!8}88.9 & \cellcolor{blue!8}90.0 & \cellcolor{red!8}91.7 & \cellcolor{red!8}95.3 & \cellcolor{red!8}89.2 & \cellcolor{red!8}81.3  \\
\bottomrule
\end{tabular}
}
\end{table*}

\begin{table*}[t]
\centering
\caption{Performance on \textsc{Search-Augmented QA}. 
We report accuracy (\%) on single-hop and multi-hop datasets.}
\label{tab:search_performance}
\small
\setlength{\tabcolsep}{6pt}
\resizebox{.9\textwidth}{!}{
\begin{tabular}{lccc|ccc|c}
\toprule
\multirow{2}{*}{\textbf{Method}} & \multicolumn{3}{c|}{\textbf{Single-Hop QA}} & \multicolumn{3}{c|}{\textbf{Multi-Hop QA}} & \multirow{2}{*}{\textbf{Avg.}} \\
 & NQ & TriviaQA & PopQA & HotpotQA & 2Wiki & MuSiQue & \\
\midrule
\rowcolor{gray!10} \multicolumn{8}{l}{Base Model: \textit{Qwen2.5-7B-Instruct}} \\
Origin & 11.6 & 35.6 & 1.20 & 16.4 & 22.2 & 4.80 & 15.3 \\
Search-R1 & 39.3 & 61.0 & 39.7 & 37.0 & 40.1 & 14.6 & 38.6 \\
ZeroSearch & 43.6 & 61.8 & \cellcolor{red!8}51.5 & 34.6 & 35.2 & 18.4 & 40.9 \\
StepSearch & 37.7 & 54.7 & 38.6 & 37.0 & 40.9 & \cellcolor{red!8}34.6 & 40.6 \\
EvolveR & 43.5 & \cellcolor{blue!8}63.4 & 44.6 & 38.2 & 42.0 & 15.6 & 41.2 \\
SkillRL(\textit{O3}) & 45.9 & 63.3 & \cellcolor{blue!8}45.9 & 43.2 & 40.3 & 20.2 & 43.1 \\
\textbf{D2Skill}(\textit{O3}) & \cellcolor{red!8}48.7 & \cellcolor{blue!8}63.4 & 44.9 & \cellcolor{red!8}47.5 & \cellcolor{blue!8}43.7 & \cellcolor{blue!8}21.0 & \cellcolor{red!8}44.9 \\
\textbf{D2Skill}(\textit{Self}) & \cellcolor{blue!8}46.7 & \cellcolor{red!8}64.0 & 42.3 & \cellcolor{blue!8}43.9 & \cellcolor{red!8}45.9 & 20.0 & \cellcolor{blue!8}43.8 \\
\midrule
\rowcolor{gray!10} \multicolumn{8}{l}{Base Model: \textit{Qwen3-4B-Instruct-2507}} \\
Origin & 18.4 & 45.5 & 28.4 & 22.8 & 18.0 & 6.3 & 23.2 \\
Search-R1 & 39.5 & \cellcolor{red!8}62.6 & 41.3 & 34.9 & 24.8 & 12.9 & 36.0 \\
Dr.Zero & 41.3 & \cellcolor{blue!8}61.6 & \cellcolor{blue!8}44.0 & 38.4 & 29.8 & 14.4 & 38.3 \\
\textbf{D2Skill}(\textit{O3}) & \cellcolor{blue!8}44.7 & 61.4 & 42.6 & \cellcolor{red!8}42.2 & \cellcolor{red!8}46.5 & \cellcolor{red!8}19.7 & \cellcolor{red!8}42.8 \\
\textbf{D2Skill}(\textit{Self}) & \cellcolor{red!8}44.9 & 60.2 & \cellcolor{red!8}44.8 & \cellcolor{blue!8}41.6 & \cellcolor{blue!8}44.7 & \cellcolor{blue!8}16.7 & \cellcolor{blue!8}42.1 \\
\bottomrule
\end{tabular}
}
\end{table*}

Tables~\ref{tab:performance} and~\ref{tab:search_performance} show that \textbf{D2Skill consistently outperforms skill-free baselines across all three benchmarks}. Across the strongest D2Skill variants in Table~\ref{tab:performance}, it reaches \textbf{90.6} overall success on \textsc{ALFWorld} and \textbf{84.4} success on \textsc{WebShop} under \textit{Qwen2.5-7B-Instruct}. The same trend holds for the smaller \textit{Qwen3-4B-Instruct-2507} \citep{yang2025qwen3}, where D2Skill improves \textsc{ALFWorld} overall success from \textbf{53.9} with GRPO to \textbf{72.7}, and remains stronger than GRPO on \textsc{WebShop}. Compared with the most relevant skill-based baseline, D2Skill also yields clear gains on \textsc{WebShop} and remains competitive or stronger on \textsc{ALFWorld}. Baseline settings are summarized in Appendix~\ref{app:baselines}.

D2Skill also transfers effectively to \textsc{Search-Augmented QA}. Under \textit{Qwen2.5-7B-Instruct}, it achieves \textbf{44.9} average accuracy and outperforms both skill-free and skill-augmented baselines overall, with especially clear gains on \textbf{multi-hop QA} such as \textsc{HotpotQA} and \textsc{2Wiki}. We view this pattern as consistent with the role of step skills: multi-hop search requires a sequence of locally correct decisions, making fine-grained guidance particularly valuable. The same trend holds for \textit{Qwen3-4B-Instruct-2507}, where average accuracy improves from \textbf{36.0} with Search-R1 to \textbf{42.8}.

We further evaluate D2Skill in the \textit{Self} setting, where skills are generated without closed-source reflectors. Even under self-feedback alone, D2Skill still improves over GRPO, reaching \textbf{82.8} overall success on \textsc{ALFWorld} and \textbf{81.3} success on \textsc{WebShop} under \textit{Qwen2.5-7B-Instruct}. The same pattern also appears for \textit{Qwen3-4B-Instruct-2507}. These results suggest that the core framework gain does not depend on privileged closed-source supervision, although stronger external reflectors can still further improve trajectory diagnosis and skill abstraction.

Finally, we evaluate D2Skill under a stronger initialization setting by starting from a teacher-initialized policy based on SFT. In this setting, D2Skill improves \textbf{both training efficiency and final performance}: after only \textbf{40} training steps, it reaches \textbf{92.2} on \textsc{ALFWorld}, already close to GRPO trained for 120 steps, and after 120 steps it further improves to \textbf{95.3} while continuing to outperform GRPO under the same budget. A similar trend also appears on \textsc{WebShop}, where D2Skill remains better than GRPO under the same training budget. Together, these results show that D2Skill is effective not only in standard RL training, but also in self-feedback and stronger-initialization regimes.

\subsection{Ablation Study}
\label{exp:ablation}

We conduct ablations on \textsc{ALFWorld} with \textsc{Qwen3-4B-Instruct-2507} to assess the contribution of each component in D2Skill. During training, we report the peak success rates of the skill and baseline groups, measured by the maximum 10-step moving average, and during validation we report the best held-out success rate. We consider six ablated variants: \textbf{\textit{(i) w/o task skills}}, removing task-level skills; \textbf{\textit{(ii) w/o step skills}}, removing step-level skills; \textbf{\textit{(iii) w/o skill management}}, disabling skill pruning and retaining all accumulated skills; \textbf{\textit{(iv) w/o baseline group}}, removing paired baseline rollouts and training with absolute rewards only; \textbf{\textit{(v) w/o utility retrieval}}, removing utility-aware ranking and using similarity-only retrieval; \textbf{\textit{(vi) w/o utility module}}, removing the utility mechanism, including baseline utility estimation and updates; and \textbf{\textit{(vii) w/o skills (GRPO)}} as a skill-free reference.

\begin{table}[t]
    \centering
    \small
    \setlength{\tabcolsep}{5pt}
    \caption{Ablation Study on \textsc{ALFWorld}.}
    \label{tab:ablation}
    \resizebox{\columnwidth}{!}{
    \begin{tabular}{lccc}
    \toprule
    \multirow{2}{*}{\textbf{Method}} & \multicolumn{2}{c}{\textbf{Train}} & \multirow{2}{*}{\textbf{Val}} \\
     & \textbf{Skill} & \textbf{Baseline} & \\
    \midrule
    \textbf{D2Skill} & \cellcolor{red!8}\textbf{70.9} & \cellcolor{red!8}\textbf{65.8} & \cellcolor{red!8}\textbf{72.7} \\
    w/o task skills & 59.1 & 53.7 & 62.7 \\
    w/o step skills & 57.8 & 55.8 & 60.2 \\
    w/o skill management & 60.0 & \cellcolor{blue!8}57.4 & 57.8 \\
    w/o baseline group & \cellcolor{blue!8}63.9 & - & \cellcolor{blue!8}68.8 \\
    w/o utility retrieval & 61.4 & 51.8 & 64.8 \\
    w/o utility module & 60.3 & - & 62.5 \\
    w/o skills (GRPO) & - & 50.6 & 53.9 \\
    \bottomrule
    \end{tabular}
    }
\end{table}

The ablation results in Table~\ref{tab:ablation} reveal three main findings. First, \textbf{removing either task skills or step skills consistently reduces performance}, indicating that both high-level task guidance and fine-grained step support are important to D2Skill. The fact that \textit{w/o step skills} remains below SkillRL should not be over-interpreted, since SkillRL additionally relies on a privileged skill bank constructed from held-out trajectories, whereas this ablation remains a training-only variant within the D2Skill framework. Second, \textbf{the larger degradation caused by removing skill management} highlights the importance of dynamic bank maintenance in discarding ineffective skills and retaining compact, high-utility knowledge for reuse. Third, removing the baseline group or utility estimation results in smaller but still clear drops, suggesting that these components primarily \textbf{enhance credit assignment and skill valuation}, thereby improving optimization and retrieval quality, rather than driving the main gains directly.

\subsection{Additional Analysis}
\label{exp:analysis}

\begin{figure}[t]
    \centering
    \includegraphics[width=\columnwidth]{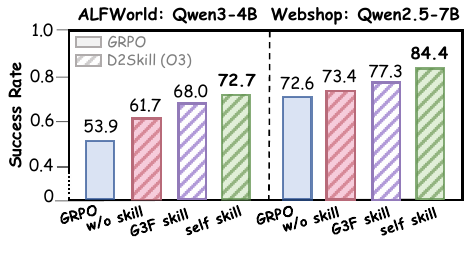}
    \caption{Eval with Different Skills.}
    \label{fig:utility_abl}
    \vspace{-0.9em}
\end{figure}

\paragraph{Utility and transferability of the skill bank.}
As shown in Figure~\ref{fig:overview_4panel_sub4}, enabling skill management yields a skill bank and retrieved skills with consistently higher average utility, indicating that utility-aware maintenance improves memory and retrieval quality by filtering ineffective skills. Figure~\ref{fig:utility_abl} further shows that the learned skills are \textbf{transferable}. Even without a skill bank at evaluation time, the policy trained with D2Skill remains competitive with, or outperforms, GRPO, suggesting that part of the gain from skill augmentation has been \textbf{internalized into the policy} during training. Moreover, using the \textit{Gemini-3-Flash}-generated skill bank from the corresponding training setting at evaluation time still yields clear gains over the no-skill variant in both \textsc{ALFWorld} and \textsc{WebShop}, while the self-generated skill bank remains the most effective. This suggests that D2Skill learns reusable skills that retain utility beyond the specific skill bank used during training.

\paragraph{Training cost.}

\begin{figure}[t]
    \centering
    \includegraphics[width=.93\columnwidth]{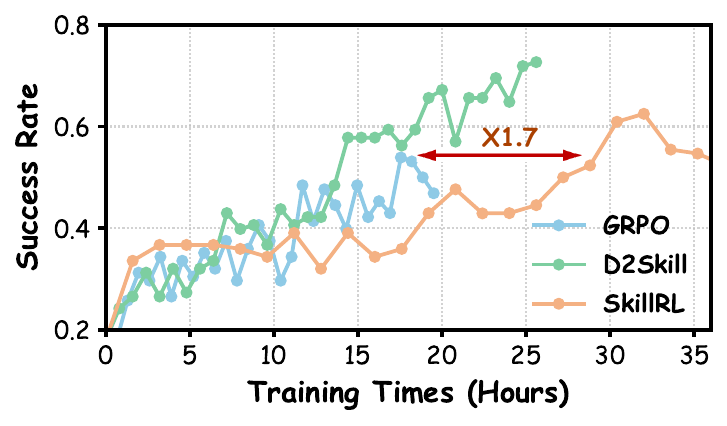}
    \caption{Val Success Dynamics.}
    \label{fig:eval_curves_times}
    \vspace{-0.9em}
\end{figure}

On \textsc{ALFWorld} with \textit{Qwen3-4B-Instruct-2507}, measured on 8$\times$H100 GPUs, D2Skill takes 25.6 wall-clock training hours, remaining \textbf{close to GRPO} (20.8 hours) while being \textbf{substantially cheaper than SkillRL} (49.2 hours). Figure~\ref{fig:eval_curves_times} further shows that D2Skill reaches strong evaluation performance much earlier in wall-clock time, making it about \textbf{1.7$\times$ faster than SkillRL} in practice. This low overhead mainly comes from an efficient retrieval pipeline: retrieval is executed with batched embedding queries and skill embeddings are updated incrementally, so only newly added skills need to be encoded after each bank update. As a result, D2Skill remains close to GRPO in training cost despite introducing dynamic skill retrieval and management. Further implementation details are available in our open-source \href{https://github.com/TU2021/D2Skill-AgenticRL}{codebase}.

\section{Related Works}

\subsection{Agent Evolution with Memory}
Recent work studies agent evolution through external memory and reusable experience, enabling LLM agents to adapt beyond parameter updates \citep{zhang2024memorysurvey,gao2025selfevolving,du2026memory}. Existing methods explore memory retention, structured updating, retrieval-aware optimization, and experience abstraction into reusable lessons, strategies, or workflows \citep{chhikara2025mem0,xu2025amem,zhou2025memento,zhao2024expel,cai2025flex}. D2Skill follows this direction, but focuses on RL-time skill valuation and maintenance through paired skill/base rollouts rather than only accumulating reusable textual memories.

\subsection{Memory-augmented Agentic RL}
Memory-augmented agentic RL treats external memory as a non-parametric complement to policy optimization, storing useful successful or failed experiences and retrieving them to guide future decisions \citep{liu2026exploratory,zhou2026memento}. Recent methods increasingly organize such experience into reusable guidance for planning and action selection, with skills emerging as a particularly effective abstraction for long-horizon agentic tasks \citep{zhai2025agentevolver,cai2025training,wang2026openclaw,xia2026skillrl}. D2Skill is most closely related to this line, but differs by jointly modeling task- and step-level skills and by dynamically valuing and managing them during training; further connections to hindsight learning, hierarchical abstraction, and contemporaneous methods are discussed in Appendix~\ref{app:related}.

\section{Conclusion}
We presented \textbf{D2Skill}, a dynamic dual-granularity skill bank framework for agentic RL that integrates task-level guidance, step-level correction, and utility-aware skill management within a joint training loop. Across interactive benchmarks and search-augmented QA, D2Skill consistently improves over strong baselines, while ablations and analyses confirm the importance of both dual-granularity skill modeling and dynamic bank maintenance.
Despite these results, our study is still limited to a small set of benchmark environments, and the strongest settings continue to rely on external reflector models. Extending D2Skill to broader agentic domains while reducing this dependence remains an important direction for future work.

\clearpage
\section*{Limitations}
We note several limitations of the current study:
\begin{itemize}[leftmargin=*, itemsep=0pt]
\item \textbf{Sensitivity of skill construction.} The quality of constructed skills still depends on the supervision source and training setup used during reflection and initialization. Improving the stability of skill construction under weaker or fully self-generated supervision remains an important direction for future work.
\item \textbf{Benchmark scope.} We evaluate D2Skill on \textsc{ALFWorld}, \textsc{WebShop}, and \textsc{Search-Augmented QA}, which cover representative interactive and search-based settings, but do not yet capture the full diversity and noise of more open-ended real-world environments.
\item \textbf{Skill quality and retrieval robustness.} The quality of generated skills and their utility estimates still depends on the underlying policy and reflector. In more complex environments, noisy, overly specific, or outdated skills may still affect retrieval quality and memory management.
\end{itemize}

\section*{Ethics Statement}
This work uses public benchmarks only; it involves no human subjects, private data, or personally identifiable information, and we anticipate no direct harmful societal impact.

\bibliography{d2skill}

\clearpage
\appendix
\section{Preliminaries}
\label{app:preliminaries}

\subsection{Agentic RL as a History-Augmented Decision Process}

We consider agentic RL in long-horizon environments modeled as a Markov decision process (MDP)
$\mathcal{M} = (\mathcal{S}, \mathcal{A}, P, r, \gamma)$,
where $s_t \in \mathcal{S}$, $a_t \in \mathcal{A}$,
and $(s_{t+1}, r_t, d_t) \sim P(\cdot \mid s_t, a_t)$.
Unlike classical RL, the policy does not directly observe the environment state.
Instead, the agent interacts through a textual interface that provides a partial description of the task and the interaction history.

For a task instance $g$, let $\tau_g$ denote the task specification,
$o_t$ the textual observation at step $t$,
and $\mathcal{H}_t^{L}=\{(o_{t-L},a_{t-L}),\dots,(o_{t-1},a_{t-1})\}$ the most recent $L$ observation--action pairs retained in the prompt.
Let $\mathcal{A}_t^{\mathrm{adm}} \subseteq \mathcal{A}$ be the admissible action set.
The policy acts on the effective context
$x_t = (\tau_g,\mathcal{H}_t^{L},o_t,\mathcal{A}_t^{\mathrm{adm}})$,
and selects actions according to $\pi_\theta(a_t \mid x_t)$.
Although the underlying dynamics are Markovian in $(s_t,a_t)$, the context $x_t$ is a fixed-window summary of past interactions and is generally not a sufficient statistic of the latent state, so the resulting MDP can be viewed as a history-augmented partially observable MDP.

\subsection{Skill Bank as an External Knowledge Store}
\label{app:skill_bank}
In addition to the sliding window $\mathcal{H}_t^{L}$, we maintain a persistent skill bank
$\mathcal{M}$,
where each skill $m\in\mathcal{M}$ stores language guidance for decision making.
At step $t$, a retrieval operator selects a small set of relevant skills
$m_t \subseteq \mathcal{M}$ conditioned on the current context
$x_t=(\tau_g,\mathcal{H}_t^{L},o_t,\mathcal{A}_t^{\mathrm{adm}})$,
and the policy acts on the augmented context
$\tilde{x}_t=(x_t,m_t)$.

Taking GRPO \citep{shao2024deepseekmath} as the RL algorithm for example.
For each task $g$, a group of $N$ trajectories is sampled and advantages are computed by normalizing returns within the group.
Under the skill-augmented context, the GRPO objective is
\(\bar r_i(\theta)=\operatorname{clip}(r_i(\theta),1-\epsilon,1+\epsilon)\):
\begin{equation*}
\begin{aligned}
\mathcal{L}_{\mathrm{GRPO}}(\theta)
& = \mathbb{E}_{i}\big[
\min\!\left(
r_i(\theta)\hat{A}_i,
\bar r_i(\theta)\hat{A}_i
\right) \\
&-\beta
D_{\mathrm{KL}}
\big(
\pi_\theta(\cdot \mid \tilde{x}_i)
\|
\pi_{\mathrm{ref}}(\cdot \mid \tilde{x}_i)
\big)
\big].
\end{aligned}
\end{equation*}
Here $\tilde{x}_t^i$ denotes the context augmented with retrieved skills, $\hat{A}_{i,t}$ denote the normalized advantage and
$r_{i,t}(\theta)=\frac{\pi_\theta(a_{i,t}\mid \tilde{x}_{i,t})}{\pi_{\theta_{\mathrm{old}}}(a_{i,t}\mid \tilde{x}_{i,t})}$
be the likelihood ratio.
The policy is optimized under skill-augmented observations, while the objective remains the same as in standard RL.

\section{Overall Training Algorithm}
\label{sec:algorithm}

For clarity, we provide the detailed training procedure of \textbf{D2Skill} in Algorithm~\ref{alg:d2skill}.  
The algorithm summarizes how \textbf{paired rollout construction}, \textbf{dual-granularity skill retrieval}, \textbf{hindsight-based optimization}, and \textbf{reflection-driven skill updating} are integrated within a unified training loop.

\begin{figure*}[t]
\refstepcounter{algorithm}
\label{alg:d2skill}
\centering
\begin{minipage}{0.98\textwidth}
\hrule
\vspace{0.5em}
\textbf{Algorithm \thealgorithm. D2Skill: Dynamic Dual-Granularity Skill Bank for Agentic RL}
\par
\vspace{0.4em}
\hrule
\vspace{0.5em}
\begin{algorithmic}[1]
\REQUIRE Policy $\pi_\theta$; task set $\mathcal{D}$; skill banks $\mathcal{M}_{\mathrm{task}}, \mathcal{M}_{\mathrm{step}}$; group size $N$; reflection threshold $\tau_{\mathrm{ref}}$
\STATE Initialize $\pi_\theta$, $\mathcal{M}_{\mathrm{task}} \leftarrow \emptyset$, $\mathcal{M}_{\mathrm{step}} \leftarrow \emptyset$
\FOR{training step $t = 1,\dots,T$}
    \STATE Sample a batch of tasks $\{g\} \sim \mathcal{D}$
    \FOR{each task $g$}
        \STATE Sample a trajectory group $\mathcal{G}_g$ with $N$ rollouts
        \STATE Split $\mathcal{G}_g$ into a \textbf{baseline group} $\mathcal{G}_g^{\mathrm{base}}$ and a \textbf{skill group} $\mathcal{G}_g^{\mathrm{skill}}$
        
        \STATE Roll out $\mathcal{G}_g^{\mathrm{base}}$ under the original context
        \[
        x_s=(\tau_g,\mathcal{H}_s^L,o_s,\mathcal{A}_s^{\mathrm{adm}})
        \]
        
        \STATE Roll out $\mathcal{G}_g^{\mathrm{skill}}$ with \textbf{dual-granularity skill retrieval}, where
        task skills are retrieved by key $g$, step skills are retrieved by key $(g,o_s)$, and injected as
        \[
        \tilde{x}_s=(x_s,m_s^{\mathrm{task}},m_s^{\mathrm{step}})
        \]
        
        \STATE Obtain trajectory returns $\{R_i\}_{i\in\mathcal{G}_g}$ and terminal success indicators $\{Y_i\}_{i\in\mathcal{G}_g}$
        \STATE Compute group success rates $\bar{Y}_g^{\mathrm{base}}$ and $\bar{Y}_g^{\mathrm{skill}}$, then form the \textbf{task-level hindsight signal}
        \[
        \Delta_g^{\mathrm{task}}=\bar{Y}_g^{\mathrm{skill}}-\bar{Y}_g^{\mathrm{base}}
        \]
        
        \FOR{each skill trajectory $i \in \mathcal{G}_g^{\mathrm{skill}}$}
            \STATE Compute the \textbf{step-level credit} and hindsight intrinsic reward
            \[
            c_i = Y_i-\bar{Y}_g^{\mathrm{base}}, \qquad
            R_i^{\mathrm{int}}=\lambda\bigl(Y_i-\bar{Y}_g^{\mathrm{base}}\bigr)
            \]
            \STATE Augment the return by $\tilde{R}_i \leftarrow R_i + R_i^{\mathrm{int}}$
            \STATE Update utilities of retrieved skills:
            \[
            \begin{aligned}
            u_m &\leftarrow (1-\beta_{\mathrm{task}})u_m+\beta_{\mathrm{task}}\Delta_g^{\mathrm{task}},\\
            u_m &\leftarrow (1-\beta_{\mathrm{step}})u_m+\beta_{\mathrm{step}}c_i
            \end{aligned}
            \]
        \ENDFOR
        
        \STATE For each $i\in\mathcal{G}_g^{\mathrm{base}}$, set $\tilde{R}_i \leftarrow R_i$
        \STATE Compute group-normalized advantages $\{A_i\}_{i\in\mathcal{G}_g}$ from $\{\tilde{R}_i\}_{i\in\mathcal{G}_g}$
        
        \IF{$\bar{Y}_g^{\mathrm{skill}} < \tau_{\mathrm{ref}}$}
            \STATE Trigger \textbf{reflection-driven skill generation} using one failed trajectory $\tau_g^{-}$ and, when available, one successful trajectory $\tau_g^{+}$
            \STATE Generate at most one task skill $m_g^{\mathrm{task}}$ and one step skill $m_g^{\mathrm{step}}$
            \STATE Deduplicate and insert them into $\mathcal{M}_{\mathrm{task}}$ and $\mathcal{M}_{\mathrm{step}}$
        \ENDIF
    \ENDFOR
    
    \STATE Update $\pi_\theta$ on all collected trajectories using GRPO with group-normalized advantages
    \STATE Periodically perform \textbf{utility-aware skill bank management} by pruning $\mathcal{M}_{\mathrm{task}}$ and $\mathcal{M}_{\mathrm{step}}$ under capacity constraints
\ENDFOR
\RETURN Trained policy $\pi_\theta$, task-skill bank $\mathcal{M}_{\mathrm{task}}$, and step-skill bank $\mathcal{M}_{\mathrm{step}}$
\end{algorithmic}
\vspace{0.3em}
\hrule
\end{minipage}
\end{figure*}

\section{Experimental Details}
\label{app:exp-details}
\subsection{Benchmarks and Evaluation Protocol}
\label{app:benchmarks}

This subsection provides the supplementary evaluation details for Table~\ref{tab:performance} unless otherwise stated. For \textsc{ALFWorld}, we report the average success rate (\%) for each subtask and the overall success rate. For \textsc{WebShop}, we report the average score and average success rate (\%).

For the two interactive benchmarks, all methods are trained for 160 training steps, evaluated every 5 training steps on 128 validation tasks by default, and reported by their best performance over the full training run unless otherwise specified. Validation is conducted on held-out tasks with the skill bank frozen, where the agent only retrieves from $\mathcal{M}_{\mathrm{task}}$ and $\mathcal{M}_{\mathrm{step}}$ without reflection or bank updates. For \textsc{Search-Augmented QA}, models are trained for 200 steps and evaluated by their test-set performance at the final 200th training step.

For model initialization, we follow SkillRL and use an SFT-initialized model for \textit{Qwen2.5-7B-Instruct} to ensure reliable instruction following under skill-augmented prompts. For \textit{Qwen3-4B-Instruct-2507}, we use the original public instruct checkpoint by default. One special case is \textsc{WebShop} with \textit{Qwen3-4B-Instruct-2507}: we find that the raw origin model almost never completes tasks successfully and yields near-zero scores regardless of subsequent RL training. Therefore, for this setting we first SFT the model on the SkillRL SFT dataset so that it acquires basic task-completion ability before RL. In addition, for each \textit{Self} setting, we further SFT the open model on about 1,000 reflection examples collected with O3 so that it has basic self-reflection ability before generating task and step skills on its own.

For the environment setup, we use a fixed seed and a short interaction history window of two observation--action pairs. Episodes are capped at 50 steps for \textsc{ALFWorld} and 15 for \textsc{WebShop}, and invalid actions are penalized during training. For \textsc{WebShop}, we adopt the small split with non-human goals. \textsc{ALFWorld} is run with the TextWorld backend \texttt{AlfredTWEnv}.

\subsection{Models and Implementation Details}
\label{app:models_impl}
This subsection summarizes the implementation details of D2Skill itself. Hyperparameters follow Table~\ref{tab:hyperparams-full}. The rollout group size is $N{=}8$, instantiating the paired baseline/skill rollout construction in Section~\ref{sec_method_1}. Retrieval hyperparameters $(\tau_{\mathrm{sim}}, m, \alpha, \eta)$ correspond to the similarity filtering and utility-aware ranking in Equation~(\ref{eq-score}), sharing the same UCB-style exploration term as the eviction rule in Equation~(\ref{eq-evict}). The number of injected skills $k$ is environment-dependent to control context length.

Reflection is triggered by $\bar{Y}_g^{\mathrm{skill}} < \tau_{\mathrm{ref}}$ (Section~\ref{sec_method_2}), and skill eviction follows Equation~(\ref{eq-evict}) under constraints $N_{\max}$ and $T_{\mathrm{prot}}$. Reflection is implemented with external LLMs (Gemini-3-Flash or O3) unless the \textit{Self} setting is used. For the stronger teacher-initialized setting, we collect 300 successful trajectories per environment using strong external agents (O3 for \textsc{ALFWorld}, Gemini-3-Pro for \textsc{WebShop}) and perform SFT before RL.

Training is conducted with distributed RL and vLLM rollouts, and reported wall-clock time corresponds to runs on 8$\times$ H100 GPUs.

\subsection{Baselines}
\label{app:baselines}
For the interactive benchmarks, we compare against three categories of baselines. \textbf{Origin} denotes the base instruct model without RL updates. \textbf{GRPO}~\citep{feng2025group} is the skill-free RL baseline. \textbf{Mem0+GRPO}~\citep{chhikara2025mem0} and \textbf{SimpleMem+GRPO}~\citep{liu2026simplemem} augment RL with external memory, while \textbf{SkillRL}~\citep{xia2026skillrl} is the most relevant prior skill-bank method. We also report direct rollout performance of strong closed-source LLMs as reference-only anchors, without RL training.

For \textsc{Search-Augmented QA}, we compare against representative search-agent baselines. \textbf{Search-R1}~\citep{jin2025searchr1} trains search-capable agents with RL. \textbf{ZeroSearch}~\citep{sun2025zerosearch} improves search behavior without requiring actual search during training. \textbf{StepSearch}~\citep{wang2025stepsearch} applies step-wise PPO to search decisions. \textbf{EvolveR}~\citep{wu2025evolver} studies experience-driven self-evolving agents, and \textbf{Dr.Zero}~\citep{yue2026drzero} develops self-evolving search agents without supervised training data. We additionally include \textbf{SkillRL}(\textit{O3}) as a skill-based baseline adapted to this setting.

\begin{table*}[t]
  \centering
  \caption{Hyperparameters and training protocol for main experiments.}
  \label{tab:hyperparams-full}
  \small
  \setlength{\tabcolsep}{4pt}
  \resizebox{\textwidth}{!}{%
  \begin{tabular}{@{}lll@{}}
    \toprule
    \textbf{Category} & \textbf{Symbol / field} & \textbf{Value} \\
    \midrule
    \multicolumn{3}{@{}l}{\emph{Benchmark / prompt context}} \\
    \quad \textsc{ALFWorld} max steps & & 50 \\
    \quad \textsc{WebShop} max steps & & 15 \\
    \quad History length $L$ & \texttt{env.history\_length} & 2 \\
    \quad Environment seed & \texttt{env.seed} & 0 \\
    \quad Invalid-action penalty coef.\ & \texttt{invalid\_action\_penalty\_coef} & 1.0 \\
    \midrule
    \multicolumn{3}{@{}l}{\emph{Batch / schedule}} \\
    \quad Training tasks per step & \texttt{train\_batch\_size} & 16 \\
    \quad Validation tasks & \texttt{val\_batch\_size} & 128 \\
    \quad Val frequency & \texttt{test\_freq} & every 5 steps \\
    \quad Training steps ${}^{*}$ & \texttt{total\_epochs} & 160 \\
    \quad Prompt max tokens & \texttt{max\_prompt\_length} & 4096 \\
    \quad Generation max tokens & \texttt{max\_response\_length} & 512 \\
    \midrule
    \multicolumn{3}{@{}l}{\emph{GRPO / PPO optimization}} \\
    \quad Group size per task & $N$ (\texttt{env.rollout.n}) & 8 \\
    \quad Skill vs.\ baseline split &  & $N/2$ each ($50\%$) \\
    \quad Learning rate &  & $1\times 10^{-6}$ \\
    \quad PPO clip range & $\epsilon$ & 0.2 \\
    \quad KL penalty coef.\ & $\beta_{\mathrm{KL}}$ & 0.01 (low-var.\ KL) \\
    \quad KL in reward &  & disabled \\
    \quad Discount & $\gamma$ & 1.0 \\
    \quad PPO epochs per update &  & 1 \\
    \quad PPO mini-batch size &  & 128 \\
    \midrule
    \multicolumn{3}{@{}l}{\emph{Rollout (vLLM)}} \\
    \quad TP size &  & 1 \\
    \quad Val sampling temperature &  & 0.4 \\
    \quad Val \texttt{do\_sample} &  & True \\
    \midrule
    \multicolumn{3}{@{}l}{\emph{D2Skill --- retrieval}} \\
    \quad Similarity threshold & $\tau_{\mathrm{sim}}$ & 0.7 \\
    \quad Stage-1 shortlist size & $m$ (\texttt{retrieval\_top\_2k}) & 10 \\
    \quad Rerank mixing weight & $\alpha$ (\texttt{retrieval\_alpha}) & 0.1 \\
    \quad UCB scale (ranking) & $\eta$ (\texttt{retrieval\_ucb\_c}) & 0.05 \\
    \quad Injected task skills $k$ (\textsc{ALFWorld}) & \texttt{top\_k\_task} & 3 \\
    \quad Injected step skills $k$ (\textsc{ALFWorld}) & \texttt{top\_k\_step} & 3 \\
    \quad Injected task / step $k$ (\textsc{WebShop}) &  & 1 / 1 \\
    \midrule
    \multicolumn{3}{@{}l}{\emph{D2Skill --- utility \& shaping}} \\
    \quad Task-skill EMA & $\beta_{\mathrm{task}}$ & 0.5 \\
    \quad Step-skill EMA & $\beta_{\mathrm{step}}$ & 0.5 \\
    \quad Hindsight intrinsic coef.\ & $\lambda$ & 1.0 \\
    \quad Credit vs.\ baseline &  & enabled \\
    \midrule
    \multicolumn{3}{@{}l}{\emph{D2Skill --- reflection \& management}} \\
    \quad Reflection threshold & $\tau_{\mathrm{ref}}$ & 0.5 (skill-group success) \\
    \quad Traj.\ cap for reflection &  & $\leq 10$ \\
    \quad Max task skills & $N_{\max}$ (\texttt{eviction\_max\_task\_skills}) & 300 \\
    \quad Max step skills & $N_{\max}$ (\texttt{eviction\_max\_step\_skills}) & 300 \\
    \quad Protection window & $T_{\mathrm{prot}}$ & 10 (training steps) \\
    \bottomrule
  \end{tabular}
  }
\end{table*}

\subsection{Prompts}
\label{app:prompts}

\paragraph{Overview.}
This section documents the prompt templates used for policy--environment interaction and reflection.
Across \textsc{ALFWorld} and \textsc{WebShop}, we keep the environment-facing prompts largely aligned with the original benchmark formats.
For skill-augmented rollouts, the only additional component is the inclusion of retrieved task-level and step-level experiences.
The templates below are presented with minimal modification to preserve their original form.

\paragraph{Policy--environment prompts.}
For both \textsc{ALFWorld} and \textsc{WebShop}, we use two layouts: a baseline layout without retrieved skills and a skill-augmented layout with retrieved experiences.
The policy prompt includes the task description, recent interaction history, current observation, and admissible actions.
The model outputs reasoning enclosed in \texttt{<think>} tags and exactly one admissible action enclosed in \texttt{<action>} tags.
These fields are parsed by the environment during rollout and evaluation.

\paragraph{Reflection prompt.}
When reflection is triggered, we construct a single prompt from one failed trajectory and, when available, one successful trajectory from the same task.
The reflector identifies the first error step and produces one step-level reflection and one task-level reflection in structured form.
These outputs are parsed and converted into skill entries for subsequent retrieval.

\begin{figure*}[t]
\centering
\begin{minipage}{0.98\textwidth}
\textbf{Prompt 1. \textsc{ALFWorld} policy template with retrieved skills.}

\vspace{0.3em}
\begin{tcolorbox}[colback=yellow!10, colframe=yellow!40!black, boxrule=0.5pt,left=1mm,right=1mm,top=0.7mm,bottom=0.7mm]
\footnotesize
\begin{Verbatim}[breaklines=true,breakanywhere=true]
You are an expert agent operating in the ALFRED embodied environment.

Your goal is to complete the following task:
{task_description}

## Current Progress
You have already taken {step_count} step(s).

Recent interaction history (observation → action):
{action_history}

Current step: {current_step}

Current observation:
{current_observation}

Admissible actions at this step:
[{admissible_actions}]

## Relevant Experience
Below are past experiences retrieved from memory. They may include:

- **Task-level experience**: for this kind of task as a whole (what to aim for, what to avoid).
- **Step-level experience**: for the current situation (what to do at this step when the observation is similar).

You should review both before deciding your next action.

When reasoning, you may:

- Use task-level experience to guide your overall plan and avoid known pitfalls
- Use step-level experience when the current observation matches the described situation
- Reuse successful actions if the situations are similar; avoid actions that previously led to failure

Warning: These lessons may be outdated. Use them only if they align with your current observation.

Retrieved experiences:
{retrieved_memories}

## Instructions
For the current step, you should follow this process:

1. Analyze the current observation
2. Review the retrieved experiences and think about whether any past experience applies
3. Reason step-by-step and choose the best admissible action

Now it's your turn to take an action.
You should first reason step-by-step about the current situation. This reasoning process MUST be enclosed within <think> </think> tags.
Once you've finished your reasoning, you should choose an admissible action for current step and present it within <action> </action> tags.
\end{Verbatim}
\end{tcolorbox}
\end{minipage}
\end{figure*}

\begin{figure*}[t]
\centering
\begin{minipage}{0.98\textwidth}
\textbf{Prompt 2. \textsc{WebShop} policy template with retrieved skills.}

\vspace{0.3em}
\begin{tcolorbox}[colback=yellow!10, colframe=yellow!40!black, boxrule=0.5pt,left=1mm,right=1mm,top=0.7mm,bottom=0.7mm]
\footnotesize
\begin{Verbatim}[breaklines=true,breakanywhere=true]
You are an expert autonomous agent operating in the WebShop e-commerce environment.

Your goal is to complete the following task:
{task_description}

## Current Progress
You have already taken {step_count} step(s).

Recent interaction history (observation → action):
{action_history}

Current step: {current_step}

Current observation:
{current_observation}

Admissible actions at this step:
[{available_actions}]

## Relevant Experience
Below are past experiences retrieved from memory. They may include:

- **Task-level experience**: for this kind of task as a whole (what to aim for, what to avoid).
- **Step-level experience**: for the current situation (what to do at this step when the observation is similar).

You should review both before deciding your next action.

When reasoning, you may:

- Use task-level experience to guide your overall plan and avoid known pitfalls
- Use step-level experience when the current observation matches the described situation
- Reuse successful actions if the situations are similar; avoid actions that previously led to failure

Warning: These lessons may be outdated. Use them only if they align with your current observation.

Retrieved experiences:
{retrieved_memories}

## Instructions
For the current step, you should follow this process:

1. Analyze the current observation
2. Review the retrieved experiences and think about whether any past experience applies
3. Reason step-by-step and choose the best admissible action

Now it's your turn to take an action.
You should first reason step-by-step about the current situation. This reasoning process MUST be enclosed within <think> </think> tags.
Once you've finished your reasoning, you should choose an admissible action for current step and present it within <action> </action> tags.
\end{Verbatim}
\end{tcolorbox}
\end{minipage}
\end{figure*}

\begin{figure*}[t]
\centering
\begin{minipage}{0.98\textwidth}
\textbf{Prompt 3. Reflection template for skill construction.}

\vspace{0.3em}
\begin{tcolorbox}[colback=yellow!10, colframe=yellow!40!black, boxrule=0.5pt,left=0.8mm,right=0.8mm,top=0.6mm,bottom=0.6mm]
\scriptsize
\begin{Verbatim}[breaklines=true,breakanywhere=true]
Output the following in order (use the exact section headers):

1) FIRST_ERROR_STEP: N
   Where N is the 1-based turn number in the *failed* trajectory where the agent first went wrong (e.g. Turn 1, Turn 2). Use 0 if unclear.

2) STEP_REFLECTION (one step-level experience for that step only):
   Output a JSON object with exactly: "title", "principle", "when_to_apply".
   This should be a concise experience for what to do at that specific step/situation (e.g. "At this step you should ...").

Example: {"title": "Check object location first", "principle": "Before picking, verify the object is in the expected receptacle.", "when_to_apply": "When the observation mentions an object but its location is unclear"}

3) TASK_REFLECTION (one task-level skill for the whole task):
   Output a JSON object with exactly: "title", "principle", "when_to_apply".
   This summarizes for the *whole task*: what to avoid and how to succeed in this kind of task (will be retrieved by task description).

Example: {"title": "Plan object location before acting", "principle": "For this task type, first identify where the object is, then plan the sequence of actions.", "when_to_apply": "When the task involves finding or moving a specific object"}

Output format (use these exact labels):
FIRST_ERROR_STEP: N

STEP_REFLECTION:
<single JSON object>

TASK_REFLECTION:
<single JSON object>
\end{Verbatim}
\end{tcolorbox}
\end{minipage}
\end{figure*}

\subsection{Examples of Skill Bank}
To illustrate the behaviors captured by the proposed dual-granularity skill bank,
we present representative examples of both \emph{task skills} and \emph{step skills}
from \textsc{ALFWorld}, \textsc{WebShop}, and \textsc{Search-Augmented QA}.

\paragraph{Task skills.}
Task skills encode reusable high-level strategies that provide global guidance throughout the trajectory, structuring exploration and mitigating systematic errors.

\paragraph{Step skills.}
Step skills encode fine-grained, context-dependent guidance at specific decision points, especially after failure, and provide targeted corrections for local mistakes. Together with task skills, they complement global planning with local refinement.

\begin{figure*}[t]
\centering
\begin{minipage}{0.98\textwidth}
\begin{tcolorbox}[colback=green!5, colframe=green!60!black, title=An Example of Task Skills (ALFWorld), boxrule=0.5pt, left=1mm, right=1mm, top=0.45mm, bottom=0.45mm]
\footnotesize
\begin{Verbatim}[breaklines=true,breakanywhere=true]
"title": "Plan the temperature-change workflow",
"principle": "For cooling tasks: 1) locate and take the target object, 2) move directly to a fridge/freezer and execute the 'cool OBJECT with FRIDGE' action, 3) carry the cooled object to the required destination and place it. Minimize wandering and avoid unrelated actions.",
"when_to_apply": "At the outset of any task that involves cooling (or heating) an object before placing it elsewhere",
"retrieval_obs": "cool some pan and put it in stoveburner."
\end{Verbatim}
\end{tcolorbox}

\vspace{0.3em}
\begin{tcolorbox}[colback=green!5, colframe=green!60!black, title=An Example of Task Skills (WebShop), boxrule=0.5pt, left=1mm, right=1mm, top=0.45mm, bottom=0.45mm]
\footnotesize
\begin{Verbatim}[breaklines=true,breakanywhere=true]
"title": "Systematically narrow down products by constraints",
"principle": "For complex find-and-buy tasks, first craft a query capturing the main attributes, then use on-page filters (color, size, price) before purchasing, verifying every required feature in the product page.",
"when_to_apply": "Whenever the task requires purchasing an item that must satisfy multiple explicit constraints (fabric, color code, size, price, care, etc.).",
"retrieval_obs": "Find me hand wash women's sweaters with long sleeve, stretch fabric, polyester spandex for teen girls, daily wear with color: xnj-tshirt345-black, and size: large, and price lower than 50.00 dollars"
\end{Verbatim}
\end{tcolorbox}

\vspace{0.3em}
\begin{tcolorbox}[colback=green!5, colframe=green!60!black, title=An Example of Task Skills (Search-Augmented QA), boxrule=0.5pt, left=1mm, right=1mm, top=0.45mm, bottom=0.45mm]
\footnotesize
\begin{Verbatim}[breaklines=true,breakanywhere=true]
"title": "Multi-hop entity verification",
"principle": "Break the question into two stages: first resolve the intermediate entity described in the query, then search for the specific attribute requested about that entity using evidence from the first hop.",
"when_to_apply": "When a search question refers to the target indirectly through another work, person, or entity.",
"retrieval_obs": "How many hardcovers appeared in the number-one position of \"The New York Times\" Bestseller List by the author of \"Night Chills\"?"
\end{Verbatim}
\end{tcolorbox}

\vspace{0.3em}
\begin{tcolorbox}[colback=blue!5, colframe=blue!60!black, title=An Example of Step Skills (ALFWorld), boxrule=0.5pt, left=1mm, right=1mm, top=0.45mm, bottom=0.45mm]
\footnotesize
\begin{Verbatim}[breaklines=true,breakanywhere=true]
"title": "Broaden the search after negative result",
"principle": "Once you verify the target object is not in the first place you looked, switch to other likely locations instead of looping over the same empty containers.",
"when_to_apply": "Right after opening the fridge and seeing it contains no potato.",
"retrieval_obs": "cool some potato and put it in garbagecan.\n\nCurrent observation: You arrive at loc 18. The cabinet 1 is open. In it, you see nothing."
\end{Verbatim}
\end{tcolorbox}

\vspace{0.3em}
\begin{tcolorbox}[colback=blue!5, colframe=blue!60!black, title=An Example of Step Skills (WebShop), boxrule=0.5pt, left=1mm, right=1mm, top=0.45mm, bottom=0.45mm]
\footnotesize
\begin{Verbatim}[breaklines=true,breakanywhere=true]
"title": "Verify option is present before selecting",
"principle": "Only issue a click when the desired option (e.g., size, color) is explicitly listed in the current observation; otherwise locate the correct menu first.",
"when_to_apply": "When the goal requires choosing a variant (size, color, etc.) and the observation does not yet display those variants.",
"retrieval_obs": "Find me loose fit, day comfort, hand wash women's tops, tees & blouses with short sleeve, polyester spandex for teen girls with color: b17-army green, and size: medium, and price lower than 40.00 dollars"
\end{Verbatim}
\end{tcolorbox}

\vspace{0.3em}
\begin{tcolorbox}[colback=blue!5, colframe=blue!60!black, title=An Example of Step Skills (Search-Augmented QA), boxrule=0.5pt, left=1mm, right=1mm, top=0.45mm, bottom=0.45mm]
\footnotesize
\begin{Verbatim}[breaklines=true,breakanywhere=true]
"title": "Identify target attributes in multi-hop questions",
"principle": "When a question links two entities, first resolve the intermediate entity, then launch a focused follow-up search for the final attribute requested by the question instead of answering from the first retrieved document.",
"when_to_apply": "When an initial search reveals the missing link in a multi-hop question but not the final target attribute.",
"retrieval_obs": "In what year was the game console for which Kowloon\'s Gate was released in 1997 released in Japan?\n\nCurrent observation: <information>{\"result\": \"Doc 1: \\\"Kowloon\'s Gate\\\" ...\"}"
\end{Verbatim}
\end{tcolorbox}
\end{minipage}
\end{figure*}

\section{Additional Related Work and Comparisons}
\label{app:related}

\subsection{Agent Evolution and Memory Management}
Recent work studies how language agents can evolve through external memory and reusable experience beyond pure parameter updates \citep{zhang2024memorysurvey,gao2025selfevolving,du2026memory}. Existing approaches examine memory retention, structured updating, retrieval-aware optimization, and experience abstraction into reusable lessons, strategies, or workflows \citep{chhikara2025mem0,xu2025amem,zhou2025memento,zhao2024expel,cai2025flex}. Relative to these methods, D2Skill focuses on \emph{RL-time} skill valuation and maintenance: skills are not only accumulated, but also scored through paired rollouts and managed throughout training.

\subsection{Memory-Augmented Agentic RL and Skill Abstraction}
Memory-augmented agentic RL treats external memory as a non-parametric complement to policy optimization, storing useful successful or failed experiences and retrieving them to guide future decisions \citep{liu2026exploratory,zhou2026memento}. More recent methods increasingly organize such experience into reusable guidance for planning and action selection, with skills emerging as a particularly effective abstraction for long-horizon agentic tasks \citep{zhai2025agentevolver,cai2025training,wang2026openclaw,xia2026skillrl}. ExpeL~\citep{zhao2024expel}, for example, distills trajectory-level textual insights and retrieves them as general reusable guidance. D2Skill is most closely related to this line, but differs from prior skill-bank methods in two key respects: it jointly models \textbf{task-level} and \textbf{step-level} skills, and it updates, values, and prunes them \textbf{during training} rather than treating the bank as largely static memory.

\subsection{Hindsight and Hierarchical Abstraction}
D2Skill is also related to hindsight learning and hierarchical abstraction in reinforcement learning. Hindsight Experience Replay reuses trajectories by relabeling goals to provide learning signals under sparse rewards \citep{andrychowicz2017hindsight}, and hindsight instruction relabeling adapts a related idea to language-model instruction following \citep{zhang2023wisdom}. Our hindsight signal is different in that it compares paired skill-injected and baseline rollouts under the same policy to estimate the utility of retrieved skills. The use of language skills as reusable abstractions is also related to hierarchical RL and language-mediated subtask abstraction \citep{jiang2019language}; D2Skill adopts a lightweight two-granularity design, separating task-level planning guidance from step-level decision support in agentic interaction.

\subsection{Comparison with Contemporaneous Work}
Contemporaneous works such as RetroAgent~\citep{zhang2026retroagent} and Complementary RL~\citep{muhtar2026complementary} are related in spirit to our approach, showing that self-evolving experience can substantially improve agentic RL performance. However, their results rely on more elaborate prompting pipelines for retrospection and experience extraction, which may increase system complexity and prompt dependence. SkillRL~\citep{xia2026skillrl} is the most closely related prior work to D2Skill. Although SkillRL also distinguishes between two task types, this mainly reflects task categorization rather than different skill granularities, and its guidance remains task-level: each task retrieves skills once and uses them throughout the trajectory. In contrast, D2Skill maintains both task skills and step skills, enabling high-level guidance and fine-grained support with retrieval at each interaction step. Moreover, D2Skill performs skill generation and management during training, rather than relying on privileged validation information for skill construction.

\end{document}